\documentclass[letterpaper]{article}
\usepackage{spconf}
\usepackage{amsmath,amssymb,amsfonts}
\usepackage{algorithmic}
\usepackage{cleveref}
\usepackage{comment}
\usepackage{graphicx,float}
\usepackage{subcaption}
\usepackage{textcomp}
\usepackage{xcolor}
\usepackage[]{algorithm}
\usepackage{bm}
\usepackage{multicol}
\usepackage{textpos}

\renewcommand\thesubsection{\Alph{subsection}}
\makeatletter

\begin{document}
\title{The Weakly-Labeled Rand Index}

\twoauthors {Dylan Stewart, Anna Hampton and Alina Zare \thanks{This material is based upon work supported by the Office of Naval Research under grant  N00014-17-1-2271.}}
{University of Florida\\
	Electrical and Computer Engineering\\
	Gainesville, FL}
{Jeff Dale and James Keller}
{University of Missouri\\
	Electrical Engineering and Computer Science\\
	Columbia, MO}
\maketitle 

\begin{abstract}
Synthetic Aperture Sonar (SAS) surveys produce imagery with large regions of transition between seabed types.  Due to these regions, it is difficult to label and segment the imagery and, furthermore, challenging to score the image segmentations appropriately. While there are many approaches to quantify performance in standard crisp segmentation schemes, drawing hard boundaries in remote sensing imagery where gradients and regions of uncertainty exist is inappropriate. These cases warrant weak labels and an associated appropriate scoring approach. In this paper, a labeling approach and associated modified version of the Rand index for weakly-labeled data is introduced to address these issues. Results are evaluated with the new index and compared to traditional segmentation evaluation methods. Experimental results on a SAS data set containing \textit{must-link} and \textit{cannot-link} labels show that our Weakly-Labeled Rand index scores segmentations appropriately in reference to qualitative performance and is more suitable than traditional quantitative metrics for scoring weakly-labeled data.
\end{abstract}

\section{Introduction}
\begin{figure*}[h]
	\centering
	\begin{subfigure}[t]{0.3\textwidth}
		\centering
		\includegraphics[width=\linewidth,height=3cm]{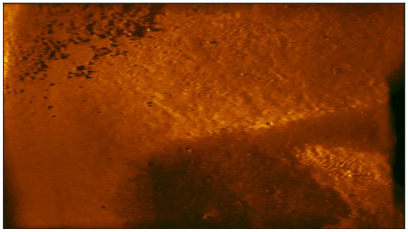}
		\caption{There are a variety of textures in SAS imagery: rocky, flat, and ripples to name a few. It would be difficult and inappropriate to assign crisp labels to this image.}	
		\label{fig:rawdataMLCL}
	\end{subfigure}
	\quad
	\begin{subfigure}[t]{0.3\textwidth}
		\centering
		\includegraphics[width=\linewidth,height=3cm]{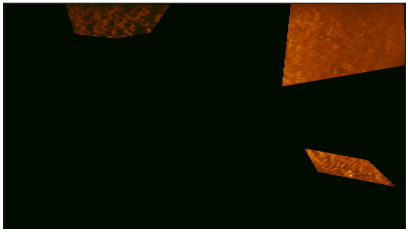}
		\caption{Three \textit{must-link} segments: rocky at the top of the image, flat at top right, and sand-ripple in the bottom right. Pixels in each highlighted segment must be in the designated segment in a proposed partitioning. }
		\label{fig:MLlabels1}	
	\end{subfigure}
	\quad
	\begin{subfigure}[t]{0.3\textwidth}
		\centering
		\includegraphics[width=\linewidth,height=3cm]{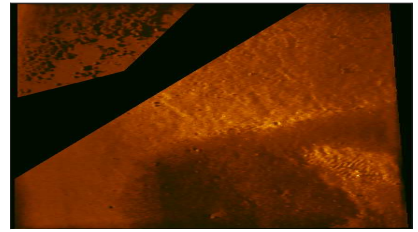}
		\caption{Two \textit{cannot-link} segments: rocky in the top left and a mixed region in the bottom right. No pixels in the rocky may be located in the same segment as pixels in the mixed region at the bottom right.}	
		\label{fig:CLlabels1}
	\end{subfigure}
	\caption{An example of our weak-labeling approach.}
    \label{fig:WLex}
\end{figure*}

In the past two decades Synthetic Aperture Sonar (SAS) systems have evolved to produce high resolution imagery of the seafloor. Descriptions of the seafloor are critical for many applications, especially mine countermeasures where awareness of environmental changes improves mine detection \cite{groen2009}. Many scene understanding methods for SAS have been developed which rely upon superpixel segmentations \cite{Zare2017Possibilistic,Peeples2019}. While metrics have been implemented to score superpixel oversegmentations in reference to crisp ground truth \cite{SPStutz16}, we develop a labeling scheme and scoring metric designed for imagery with soft boundaries and complex transitions between textures and compare our approach to traditional crisp methods. An example of this complex imagery is shown in Figure \ref{fig:rawdataMLCL} where there are no clear boundaries between the rocky portion in the top left, flat in the middle, and portion of sand-ripple in the bottom right.

There are numerous metrics that compare crisp segmentations \cite{jaccard,RANDindex,HubertAR,MartinGCE}. Jaccard \cite{jaccard} proposed a simple intersection over union (IoU). Rand \cite{RANDindex} proposed the Rand index (RI) as a method to compare multiple partitionings of a data set. The RI for a pair of partitionings is shown in Equation \ref{eq:randIndex}:
\begin{equation}
    \label{eq:randIndex}
    \text{Rand} = \frac{a+b}{a+b+c+d},
\end{equation}
where $a$ counts the pairs of points in two partitions that are in the same segment in each partition, $b$ counts the pairs that are in different segments in both partitions, and $c$ and $d$ count the pairs of points that are in the same segment in one partition but not in the other for each ordering. 

\section{WL-Rand}
\label{sec:Success}

In this section, we propose a weakly labeled Rand (WL-Rand).  The WL-Rand redefines the a, b, c and d terms in Equation \ref{eq:randIndex} to be appropriate for imagery with smooth transition regions between segments of interest. Consider an image partitioning, $\mathcal{S} = \Pi_{k=1}^K \bigcup \mathcal{S}_k$, where $\mathcal{S}_k$ is the set of location indices of the kth segment with $n_k$ pixels. 

We propose, for images with smooth transition regions, that one label pure regions as \textit{must-link} segments. Pairs of pixels that satisfy these \textit{must-link} segments in a candidate partition correspond to term $a$ in Equation \ref{eq:randIndex}. The \textit{must-link} segments, $\mathcal{M} = \Pi_{l=1}^L \bigcup \mathcal{M}_l$, are sets of pixels that must be grouped together in a given partition of the image. The correct \textit{must-link} pairs for a partitioning are: 
\begin{equation}
\label{eq:cML}
a = \sum\limits_{k=1}^K\sum\limits_{l=1}^L {\big|\mathcal{S}_k\cap \mathcal{M}_l\big| \choose 2}.
\end{equation}
This term is largest when \textit{must-link} segments are represented by few segments and smallest when the image is oversegmented. 

The \textit{cannot-link} segments, $\mathcal{C} = \Pi_{u=1}^U \bigcup \mathcal{C}_u$, are sets of pixels that should be in distinct segments. Pairs of pixels that satisfy these \textit{cannot-link} segments in a candidate partition correspond to term $b$ in Equation \ref{eq:randIndex}. It is unnecessary for pixels in a \textit{cannot-link} segment to exist in the same \textit{must-link} segment; however, any pair of pixels from disjoint \textit{cannot-link} segments must not be in the same segment. The correct \textit{cannot-link} pairs are shown in Equation \ref{eq:cCL}:
\begin{equation}
\label{eq:cCL}
b = \sum\limits_{k=1}^K\sum\limits_{u=1}^U\sum\limits_{u^\prime \neq u} \big|\mathcal{C}_{u^\prime}\big|\cdot\big|\mathcal{S}_k \cap \mathcal{C}_u \big|.
\end{equation}

The correct \textit{cannot-link} pairs for a partitioning are computed by summing all the correct pairs from each segment. For a single segment, the correct \textit{cannot-link} pairs are the number of pairs of the kth segment that are within a single \textit{cannot-link} segment. The number of correctly preserved pairs is assessed by taking the product of the size of each other \textit{cannot-link} segment with the number of pixels within the particular \textit{cannot-link} segment. This term is large when all segments adhere to strict boundaries and is small when segments cross margins. 

Incorrect \textit{must-link} pairs for a segment are computed by counting the pairs of pixels within a segment that are not contained in the same \textit{must-link} segment as adjacent pixels with the same partition label. The number of incorrect \textit{must-link} pairs, which is parallel to term $c$ in Equation \ref{eq:randIndex} are:
\begin{equation}
\label{eq:iML}
c = \sum\limits_{k=1}^K\sum\limits_{l=1}^L \big| \mathcal{M}_l\setminus \big(\mathcal{S}_k \cap \mathcal{M}_l\big) \big| \cdot \big| \mathcal{S}_k \cap \mathcal{M}_l \big | .
\end{equation}
 In the first term of Equation \ref{eq:iML}, the number of missing links are measured by the set difference between a \textit{must-link} segment and pixels within a segment that intersect the \textit{must-link} segment. The pairs that disagree with the labeling are computed by multiplying the number of missing links by the size of the pixels that are in the \textit{must-link} segment. 

This incorrect term is large if there are multiple segments that contain parts of a \textit{must-link} segment because every pair of pixels in a segment that intersects a \textit{must-link} segment should be linked together within the same \textit{must-link} segment. This term is small for a given segmentation if each \textit{must-link} segment is encompassed by a single segment. 

Lastly, pairs that are incorrect with reference to the \textit{cannot-link} segments are pairs of pixels which are found in the same segment but are intended to be in disjoint \textit{cannot-link} segments. The number of incorrect \textit{cannot-link} pairs, like term $d$ in Equation \ref{eq:randIndex}, are represented by Equation \ref{eq:iCL}:
\begin{equation}
\label{eq:iCL}
d = \sum\limits_{k=1}^K\sum\limits_{u=1}^U\sum\limits_{u^\prime \neq u} \big|\mathcal{S}_k \cap \mathcal{C}_u \big| \cdot \big| \mathcal{S}_k \cap \mathcal{C}_{u^\prime}\big|.
\end{equation}

This term is large when there are segments which cross boundaries and small otherwise. We combine these four terms to make $\text{WL-Rand} = (a+b)/(a+b+c+d)$. The index scores a segmentation in range of $[0,1]$, where $0$ would contain the worst and $1$ is the optimum in reference to a desired weakly-labeled partitioning. We experiment with WL-Rand and compare to traditional crisp methods of scoring segmentation.

\section{Experiments}
\label{sec:experiments}
We compare WL-Rand with traditional quantitative indices to score segmentations produced by hierarchical superpixel merging. Initial superpixels are fixed and an iterative merging scheme produces candidate partitionings. An adjacency matrix is computed to indicate neighboring segments. A comparison function is applied between pairs of neighboring segments where entries indicate the similarity or dissimilarity score for each segment pair. Each comparison function is applied between high-dimensional exemplars of pairs of neighboring superpixels. The most similar pair of segments are merged and the adjacency matrix, and similarity or dissimilarity matrix are updated respectively. This process continues until one distinct segment remains. Candidate partitions are produced by merging the most similar pair of superpixels and the comparison measures are updated for the neighborhood surrounding the merged pair. Each candidate partition is compared to the desired labels at each iteration and evaluated using the quantitative metrics. The desired labels for each image are either \textit{must-link} and \textit{cannot-link} labels or one of the two sets of crisp labels that were produced for the data set. We compare our labeling and scoring approach with traditional crisp labels and metrics and show qualitative results from this experiment.

Before computing the initial segments, images are downsampled to reduce dimensionality. Normalized Cuts is used to generate the initial segment segmentation, with 30 segments, as described in \cite{cobb2014boundary}. Two feature sets are computed: multi-scale Lacunarity \cite{lacunarity} combined with Sobel \cite{sobel} and pre-trained convolutional filter responses \cite{AlexNet}.

Consider $\bm{X}=\{{\bm{X}^{1},\bm{X}^{2},...,\bm{X}^{S}}\}$ the set of all $S$ segments encompassing $N$ pixels. Each segment $\bm{X}^{s}\in \mathbb{R}^{D\times n_{s}}$, contains $n_{s}$ exemplars of $D$-dimensional features. The sparse adjacency matrix, $\bm{A}\in \mathbb{R}^{S\times S}$, indicates at each entry, $a_{ij}$, if the ith and jth segments are neighbors. Consider a comparison function, $\psi$, applied to neighboring segments:

\begin{equation}
\label{eq:psi}
\Psi_{ij} = \begin{cases}
\psi(S_i,S_j) &\text{if } a_{ij}\\
\beta &\text{elsewhere},
\end{cases} 
\end{equation}
When using a similarity metric, $\bm{\Psi}\in \mathbb{R}^{S \times S}$ contains similarity scores of neighbors with a lower bound of $\beta$. Analogously, for distance metrics, each entry is a distance measure and $\beta$ becomes an upper limit. We use similarity measures with lower bounds for comparisons of mutual information measures and dissimilarity measures with upper bounds for comparisons of distances. 

\begin{figure*}
	\centering
\begin{subfigure}[t]{.105\textwidth}
	\centering
	\includegraphics[trim={0cm 0cm 0cm 0cm},clip,width=\linewidth]{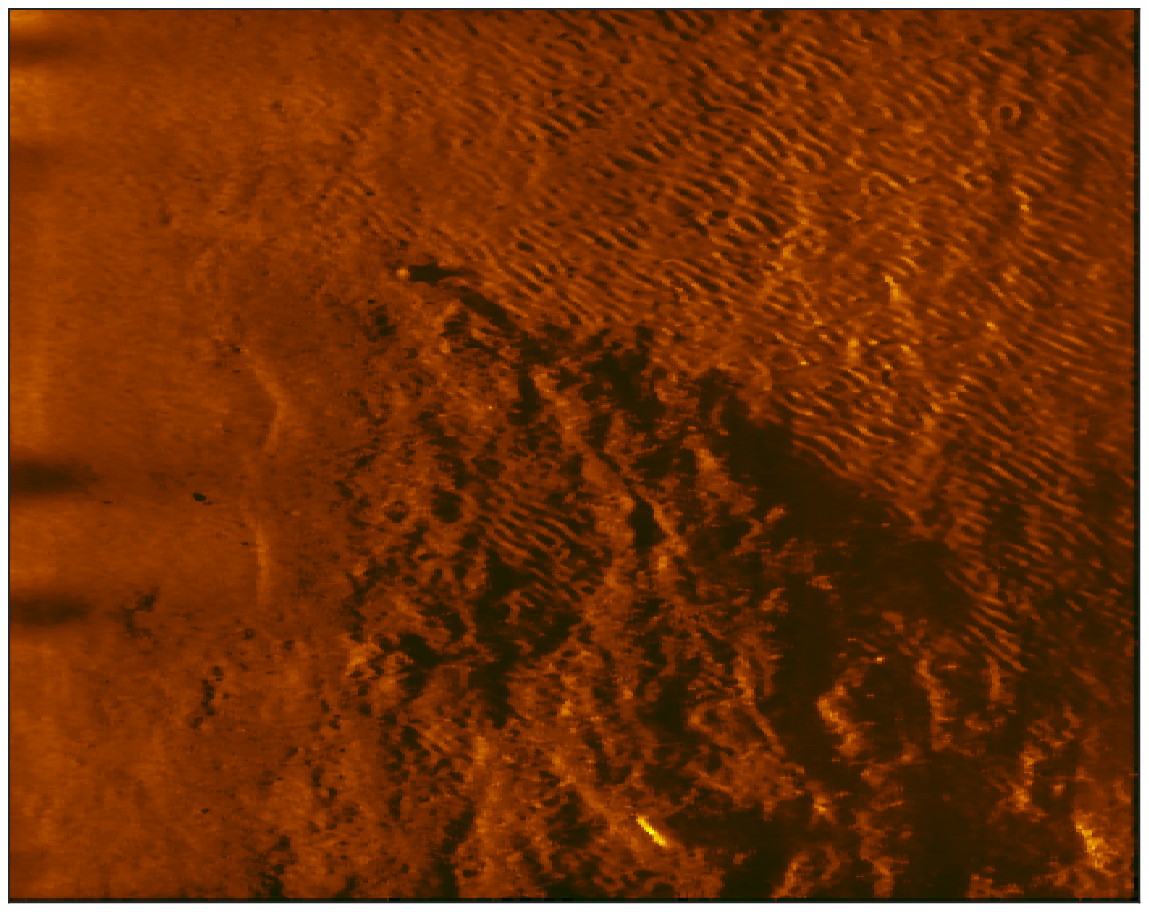}
\end{subfigure}
\begin{subfigure}[t]{.105\textwidth}
	\centering
	\includegraphics[trim={0cm 0cm 0cm 0cm},clip,width=\linewidth]{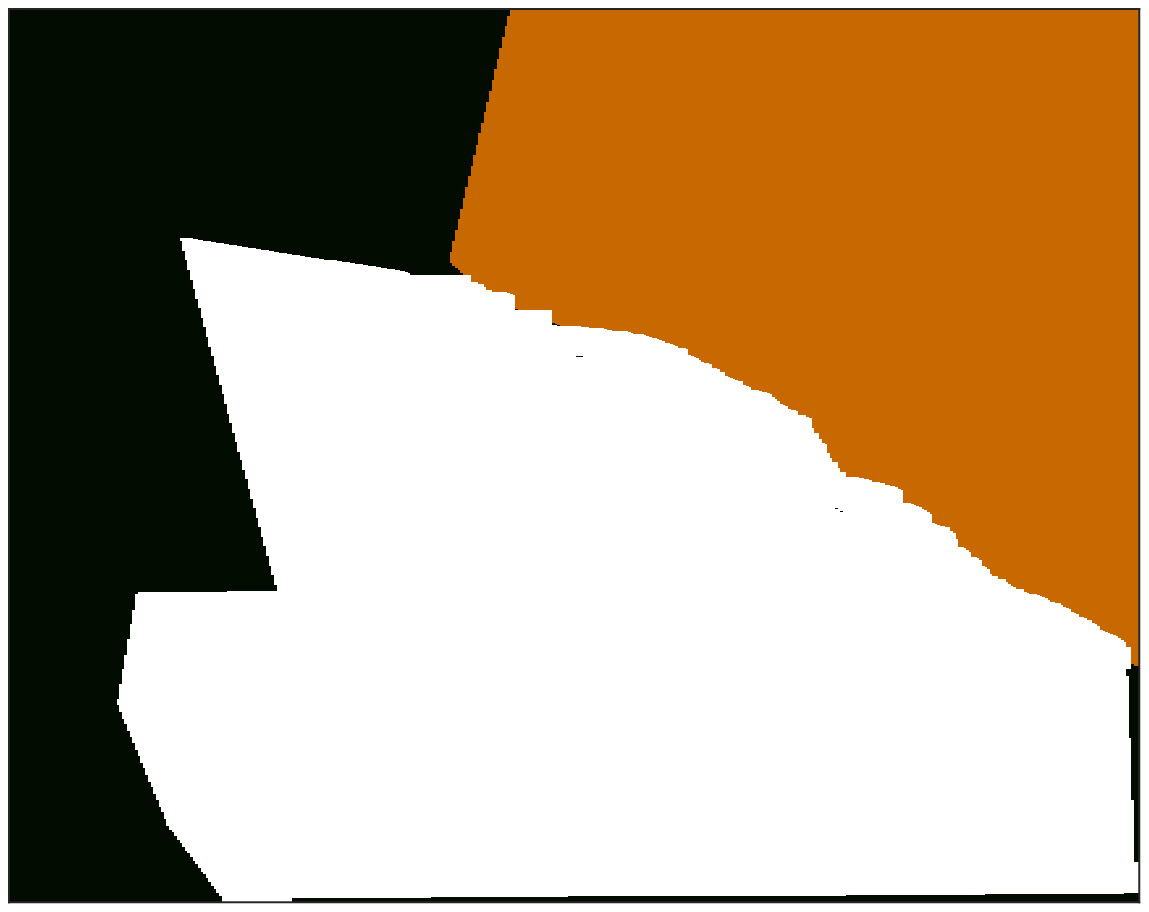}
\end{subfigure}
\begin{subfigure}[t]{.105\textwidth}
	\centering
	\includegraphics[trim={0cm 0cm 0cm 0cm},clip,width=\linewidth]{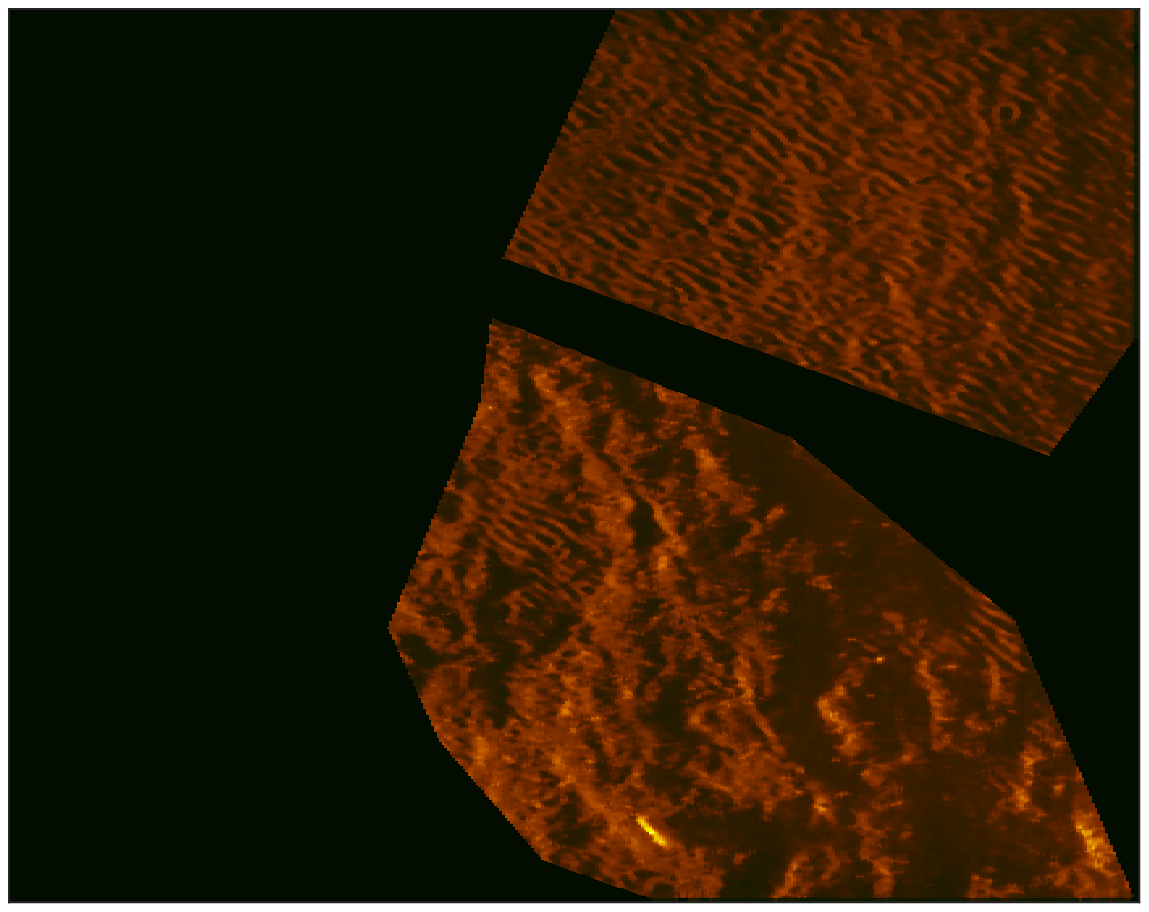}
\end{subfigure}
\begin{subfigure}[t]{.105\textwidth}
	\centering
	\includegraphics[trim={0cm 0cm 0cm 0cm},clip,width=\linewidth]{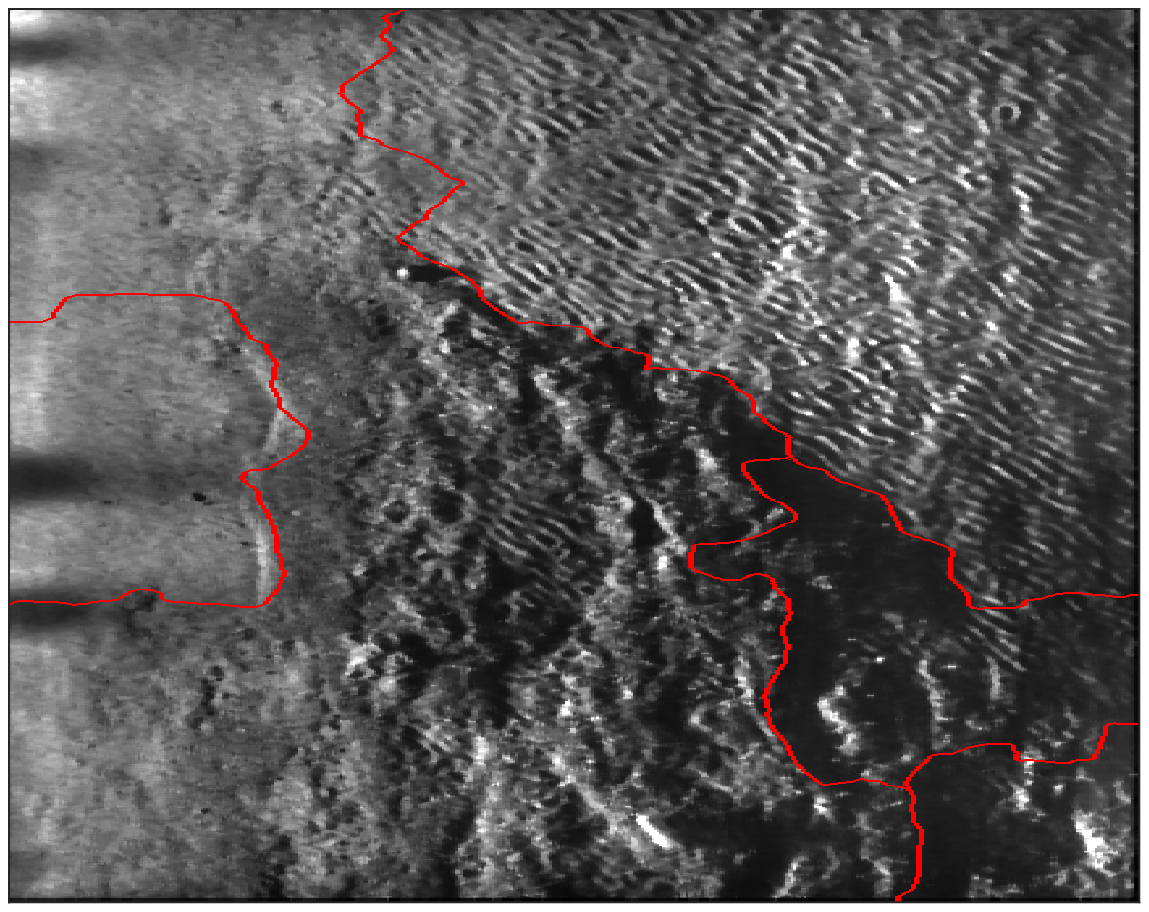}
\end{subfigure}
\begin{subfigure}[t]{.105\textwidth}
	\centering
	\includegraphics[trim={0cm 0cm 0cm 0cm},clip,width=\linewidth]{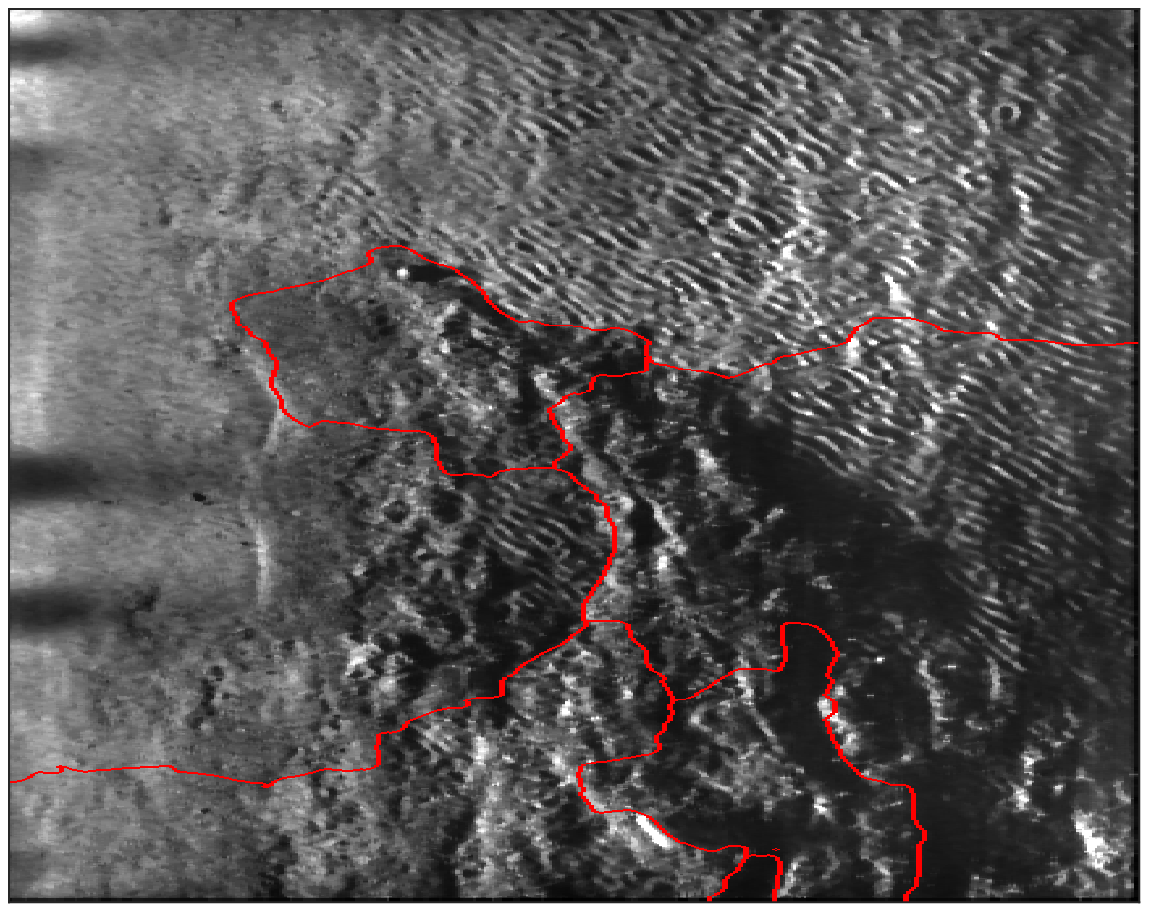}
\end{subfigure}
\begin{subfigure}[t]{.105\textwidth}
	\centering
	\includegraphics[trim={0cm 0cm 0cm 0cm},clip,width=\linewidth]{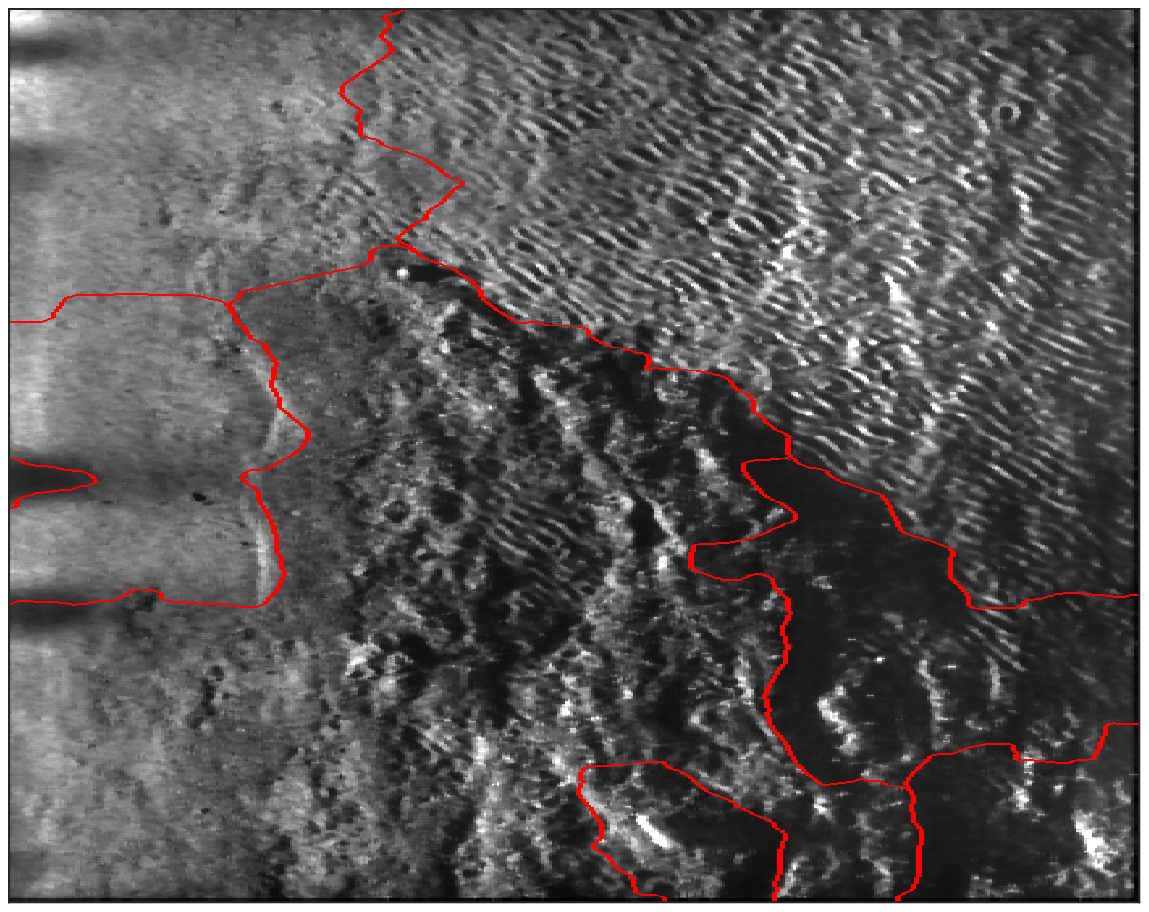}
\end{subfigure}
\begin{subfigure}[t]{.105\textwidth}
	\centering
	\includegraphics[trim={0cm 0cm 0cm 0cm},clip,width=\linewidth]{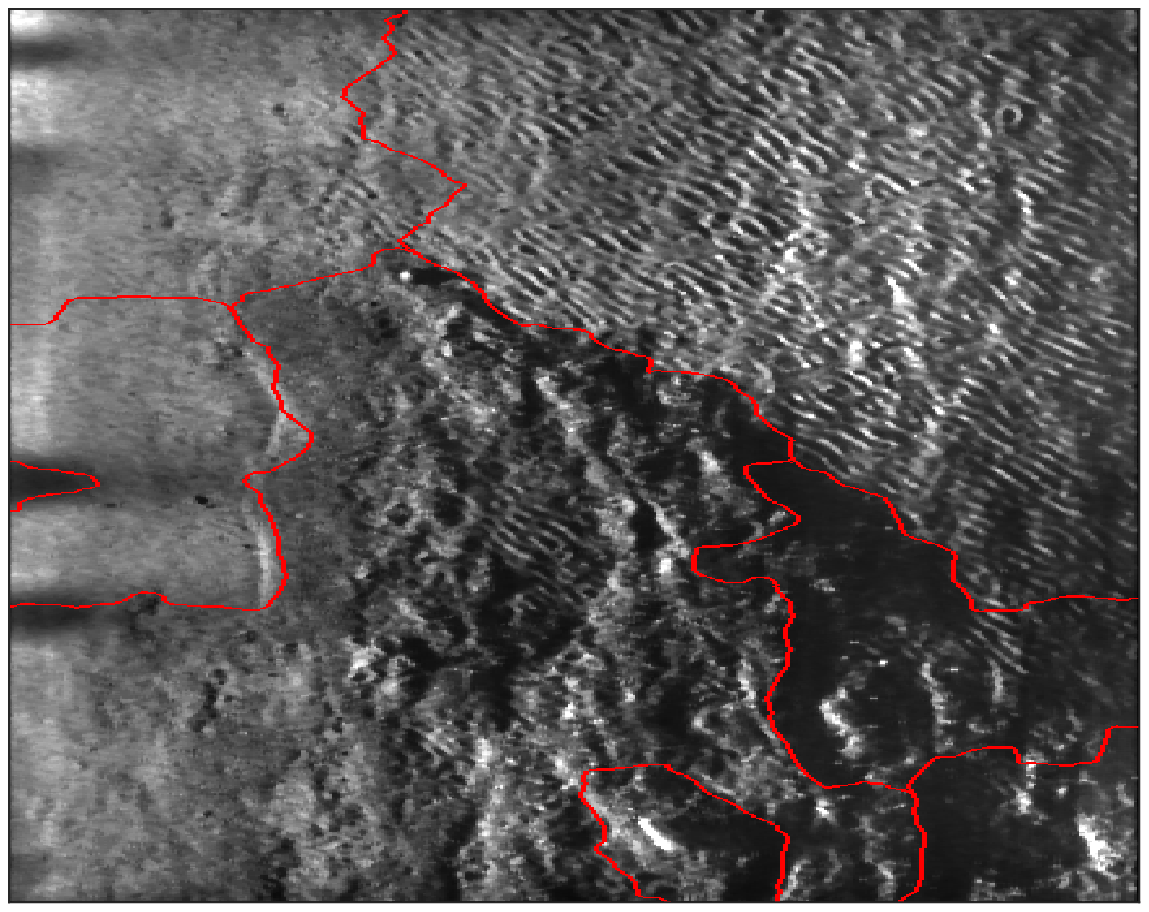}
\end{subfigure}
\begin{subfigure}[t]{.105\textwidth}
	\centering
	\includegraphics[trim={0cm 0cm 0cm 0cm},clip,width=\linewidth]{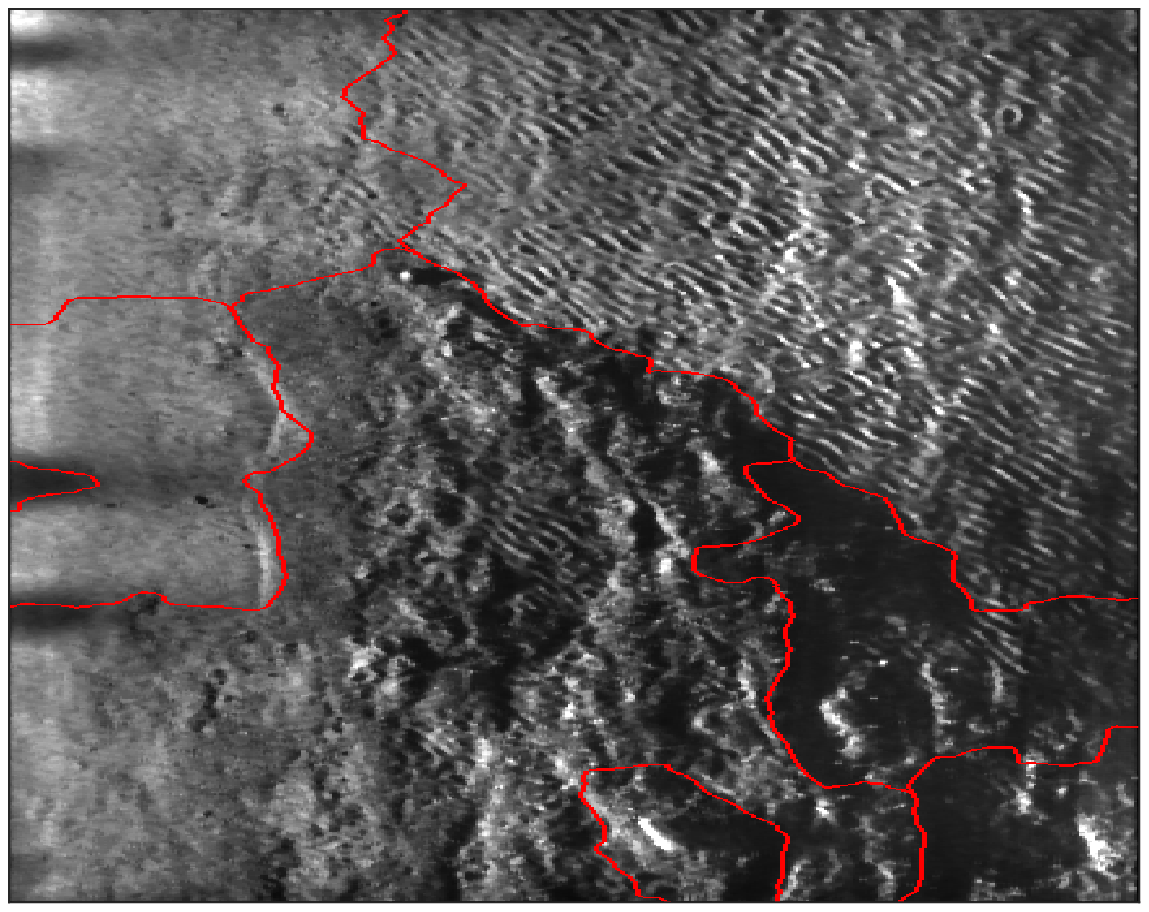}
\end{subfigure}
\begin{subfigure}[t]{.105\textwidth}
	\centering
	\includegraphics[trim={0cm 0cm 0cm 	0cm},clip,width=\linewidth]{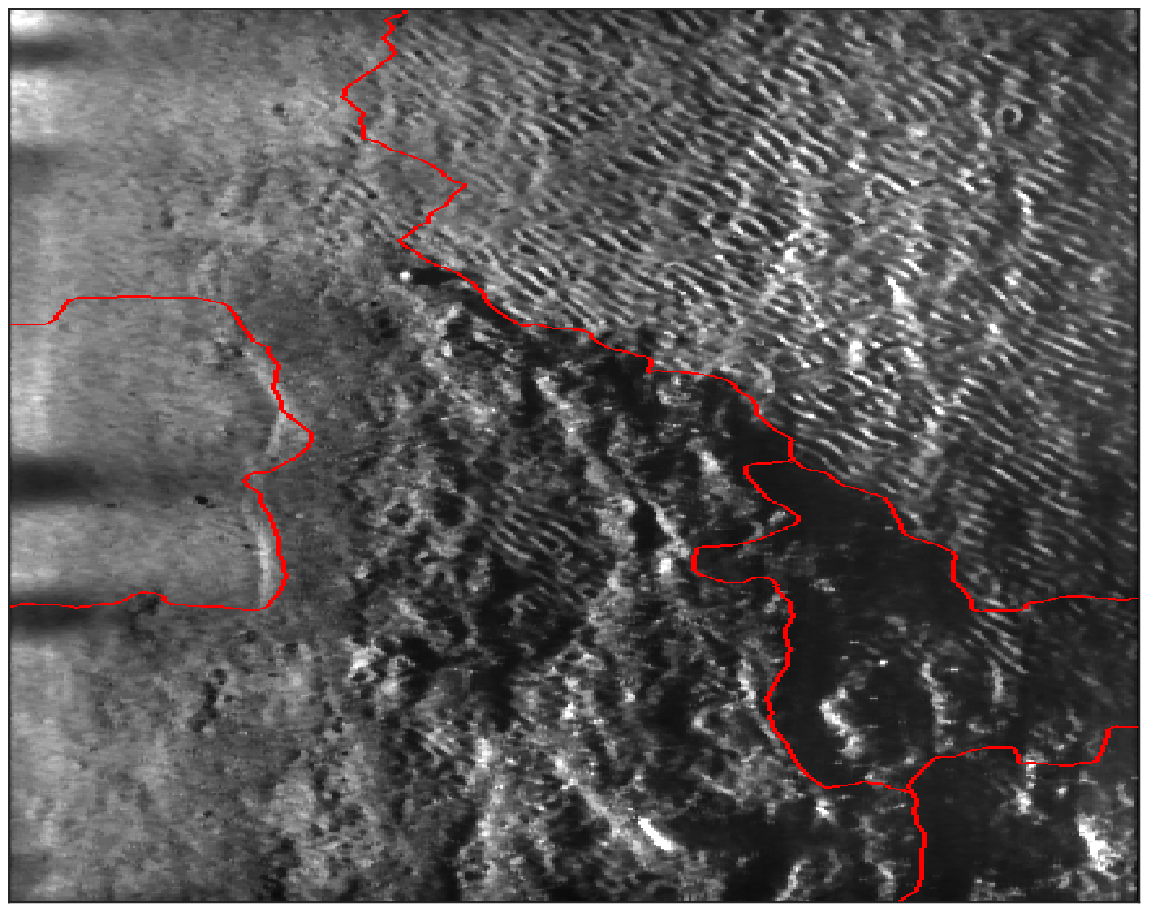}
\end{subfigure} \\

\begin{subfigure}[t]{.105\textwidth}
	\centering
	\includegraphics[trim={0cm 0cm 0cm 0cm},clip,width=\linewidth]{Figures/data/image4.eps}
\end{subfigure}
\begin{subfigure}[t]{.105\textwidth}
	\centering
	\includegraphics[trim={0cm 0cm 0cm 0cm},clip,width=\linewidth]{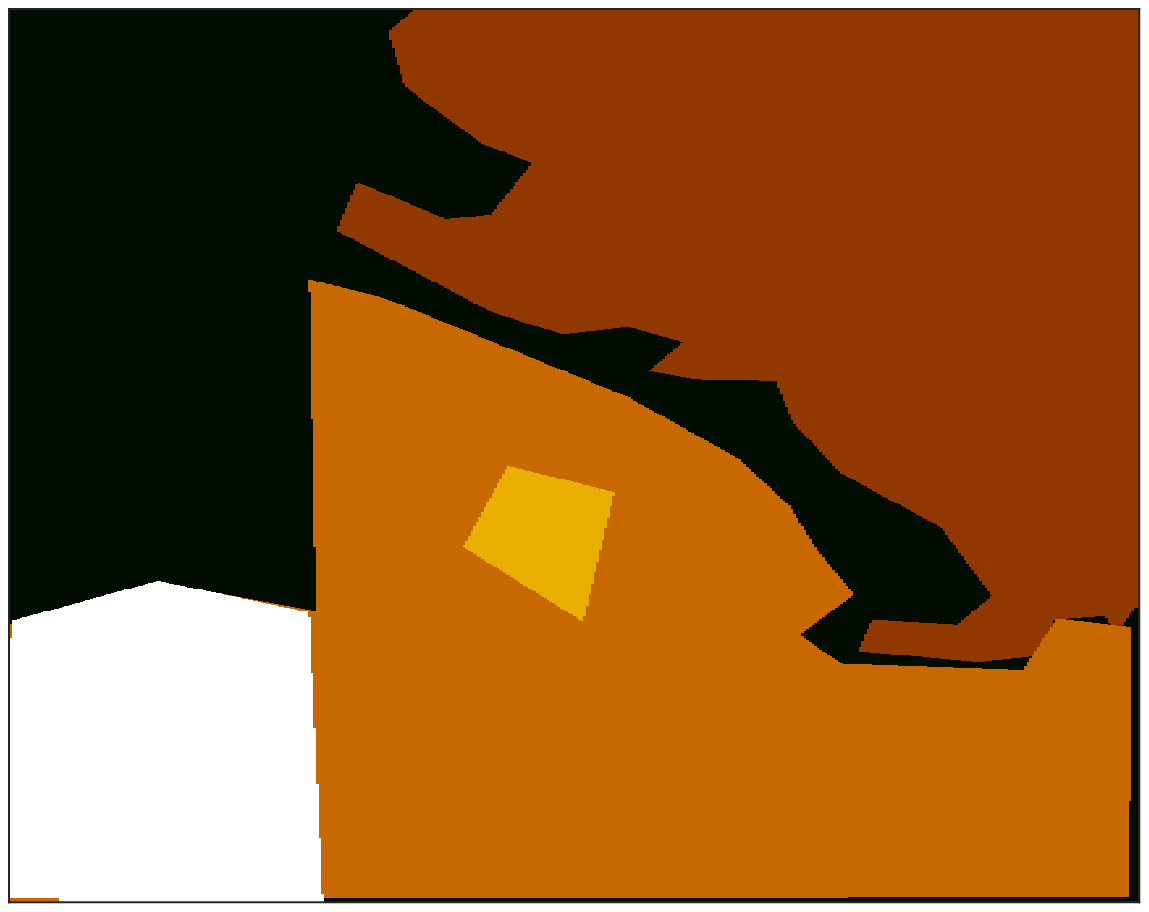}
\end{subfigure}
\begin{subfigure}[t]{.105\textwidth}
	\centering
	\includegraphics[trim={0cm 0cm 0cm 0cm},clip,width=\linewidth]{Figures/data/image4CL.eps}
\end{subfigure}
	\begin{subfigure}[t]{.105\textwidth}
	\centering
	\includegraphics[trim={0cm 0cm 0cm 0cm},clip,width=\linewidth]{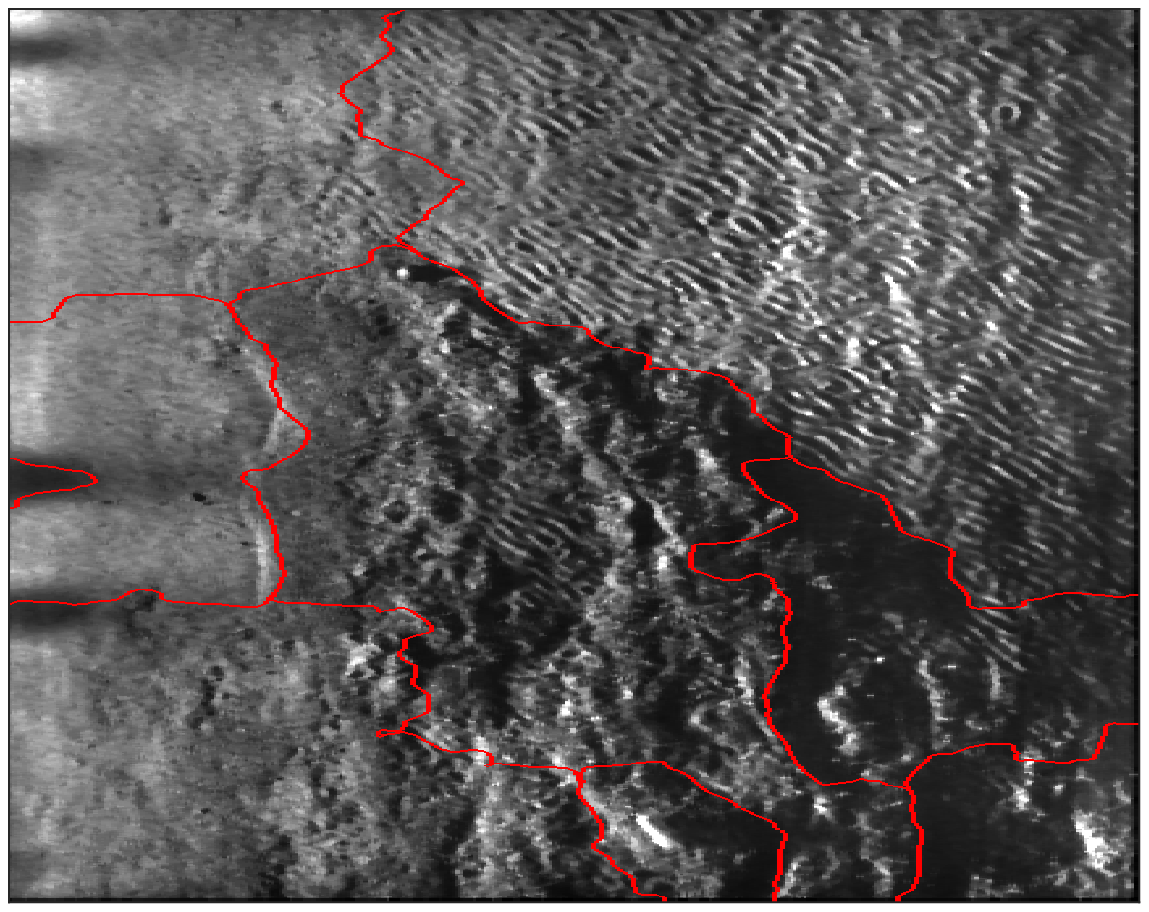}
\end{subfigure}
\begin{subfigure}[t]{.105\textwidth}
	\centering
	\includegraphics[trim={0cm 0cm 0cm 0cm},clip,width=\linewidth]{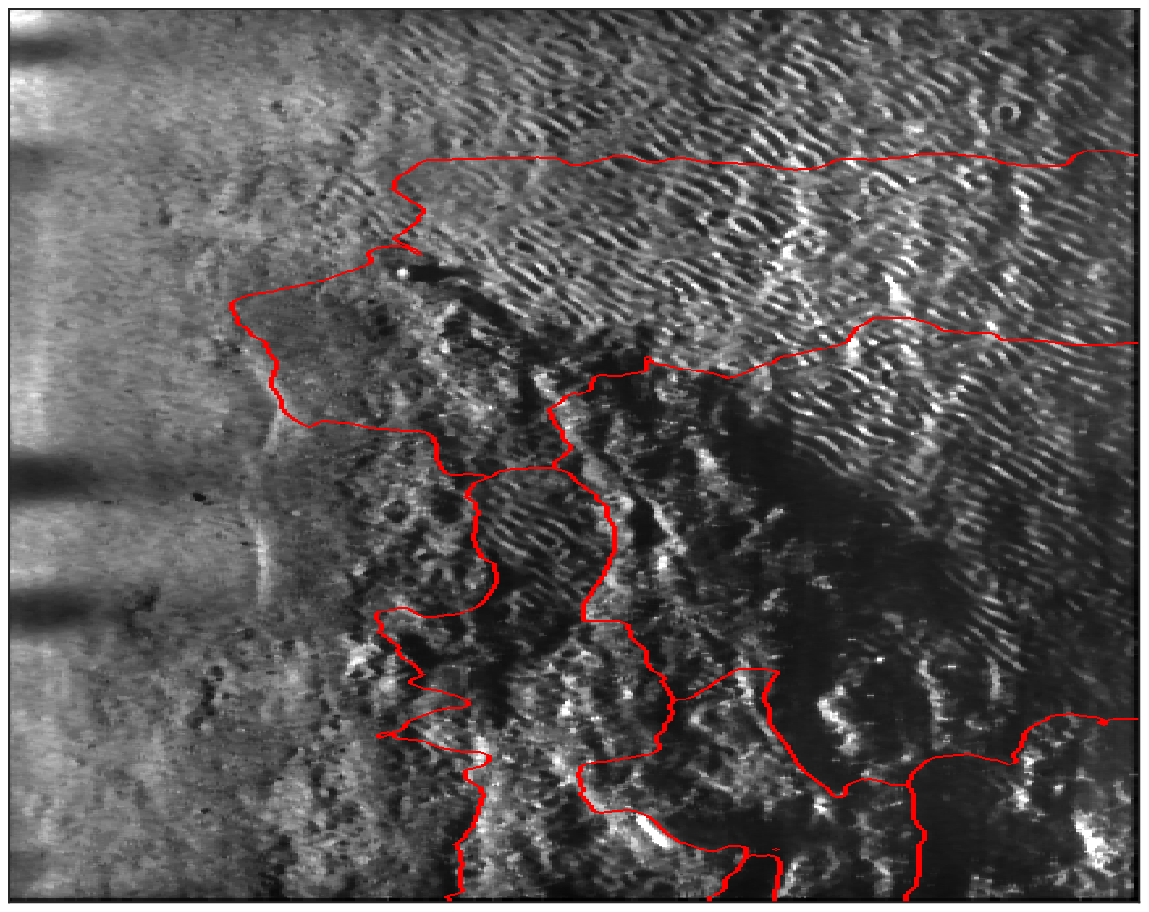}
\end{subfigure}
\begin{subfigure}[t]{.105\textwidth}
	\centering
	\includegraphics[trim={0cm 0cm 0cm 0cm},clip,width=\linewidth]{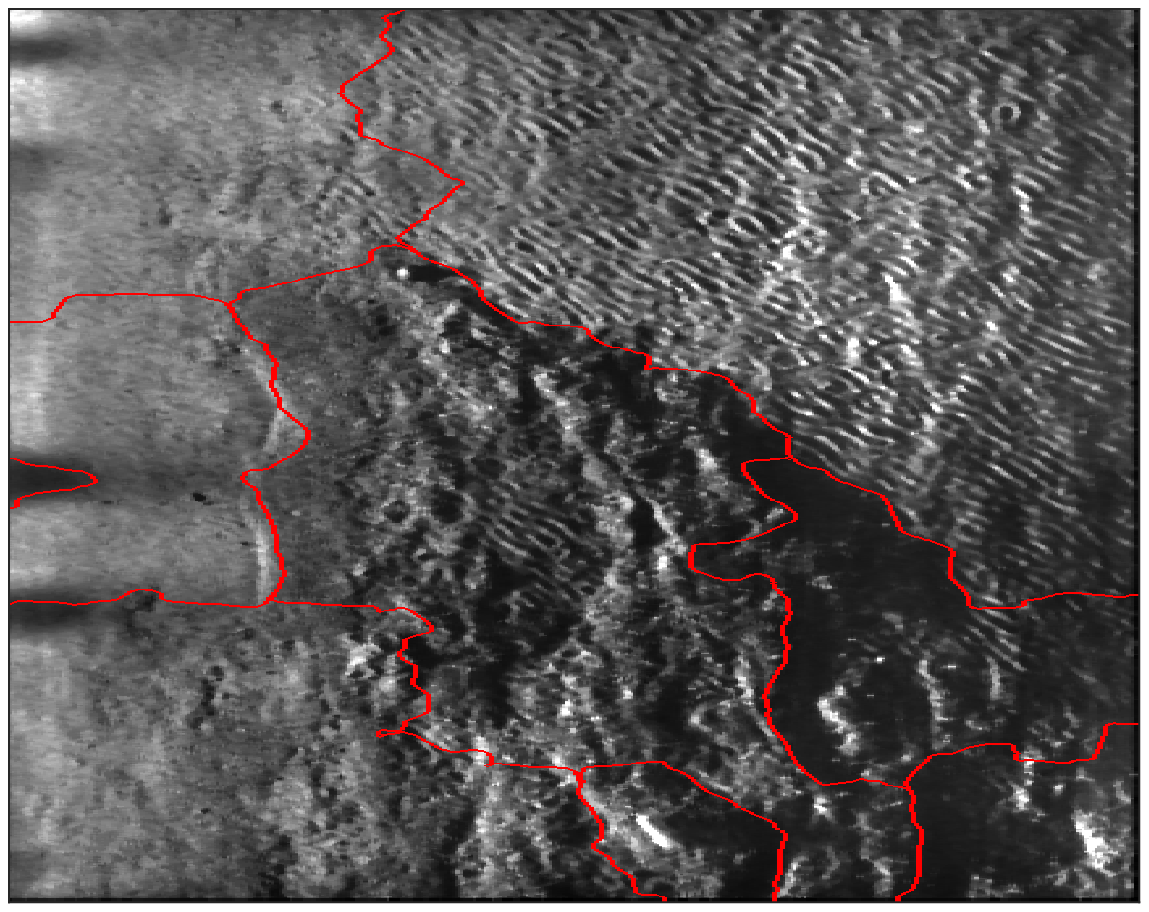}
\end{subfigure}
\begin{subfigure}[t]{.105\textwidth}
	\centering
	\includegraphics[trim={0cm 0cm 0cm 0cm},clip,width=\linewidth]{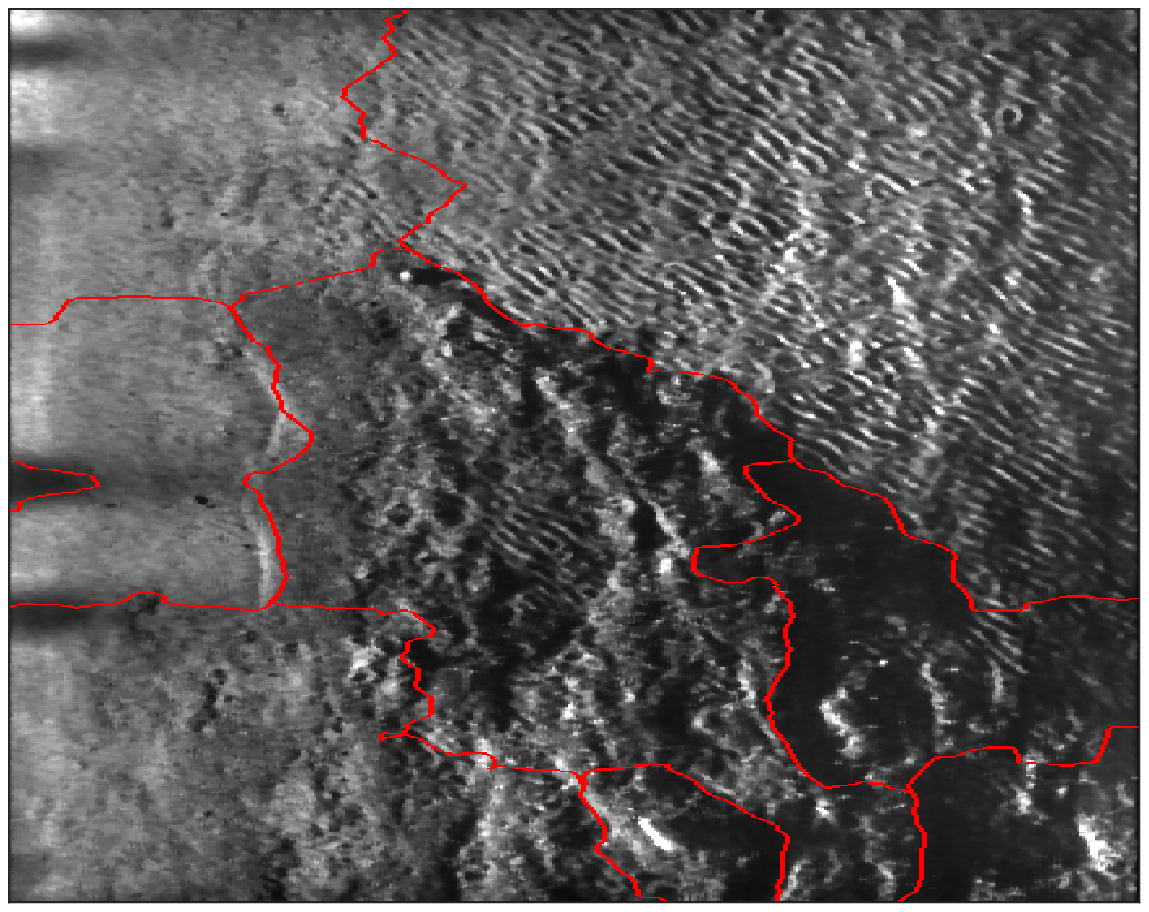}
\end{subfigure}
\begin{subfigure}[t]{.105\textwidth}
	\centering
	\includegraphics[trim={0cm 0cm 0cm 0cm},clip,width=\linewidth]{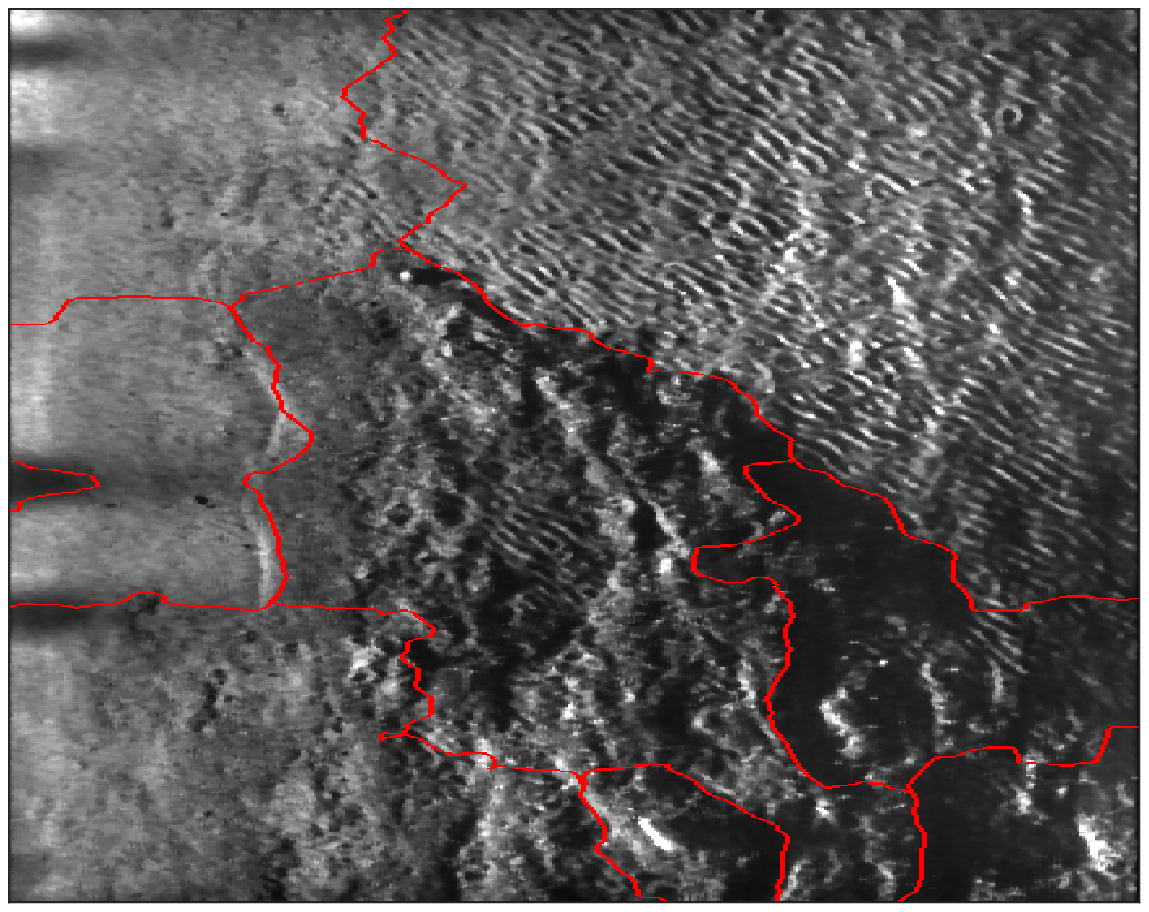}
\end{subfigure}
\begin{subfigure}[t]{.105\textwidth}
	\centering
	\includegraphics[trim={0cm 0cm 0cm 	0cm},clip,width=\linewidth]{Figures/Merging/WL-Rand/s2/PHi00-15Oct21_1246-00000006Edist.eps}
\end{subfigure}\\

	\centering
\begin{subfigure}[t]{.105\textwidth}
	\centering
	\includegraphics[trim={0cm 0cm 0cm 0cm},clip,width=\linewidth]{Figures/data/image4.eps}
	\caption{Image}
\end{subfigure}
\begin{subfigure}[t]{.105\textwidth}
	\centering
	\includegraphics[trim={0cm 0cm 0cm 0cm},clip,width=\linewidth]{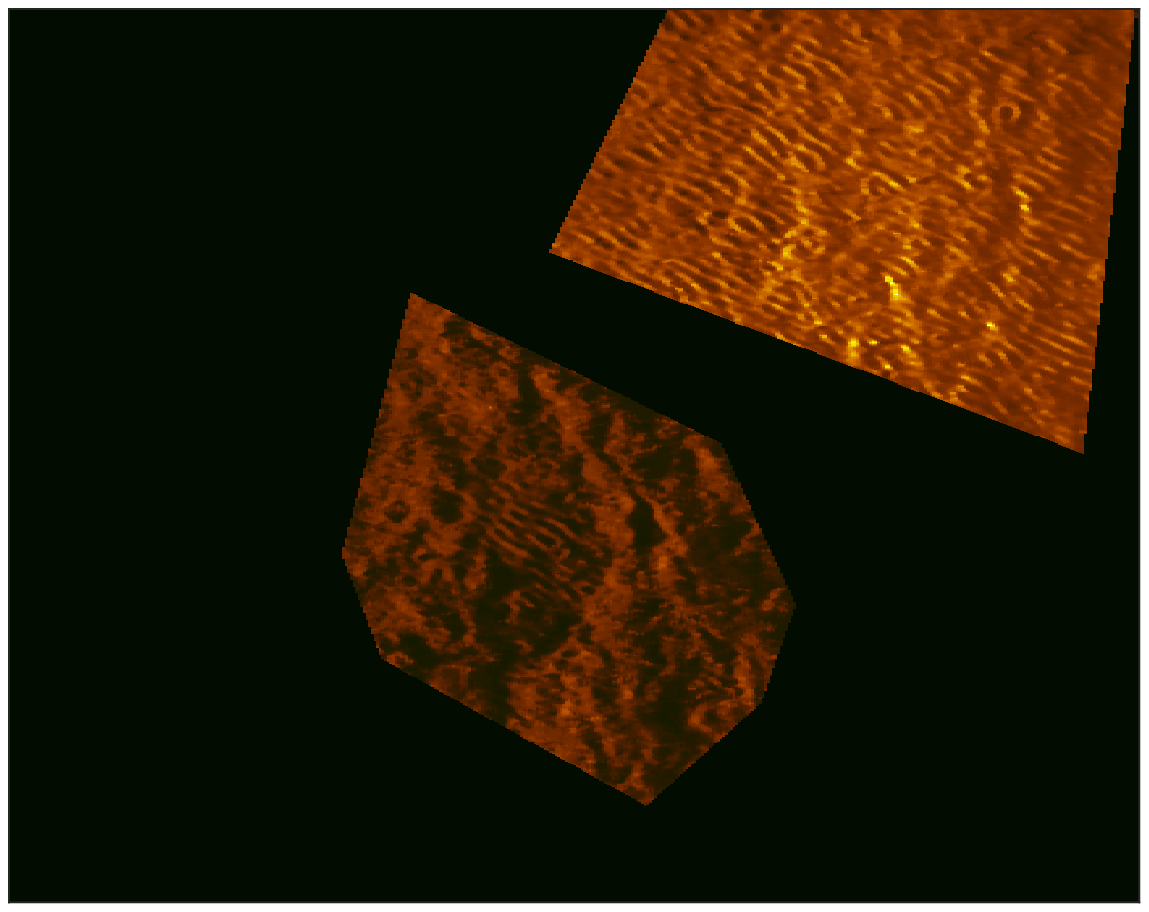}
	\caption{ML}
	\label{fig:ML}
\end{subfigure}
\begin{subfigure}[t]{.105\textwidth}
	\centering
	\includegraphics[trim={0cm 0cm 0cm 0cm},clip,width=\linewidth]{Figures/data/image4CL.eps}
	\caption{CL}
	\label{fig:CL}
\end{subfigure}	
\begin{subfigure}[t]{.105\textwidth}
	\centering
	\includegraphics[trim={0cm 0cm 0cm 0cm},clip,width=\linewidth]{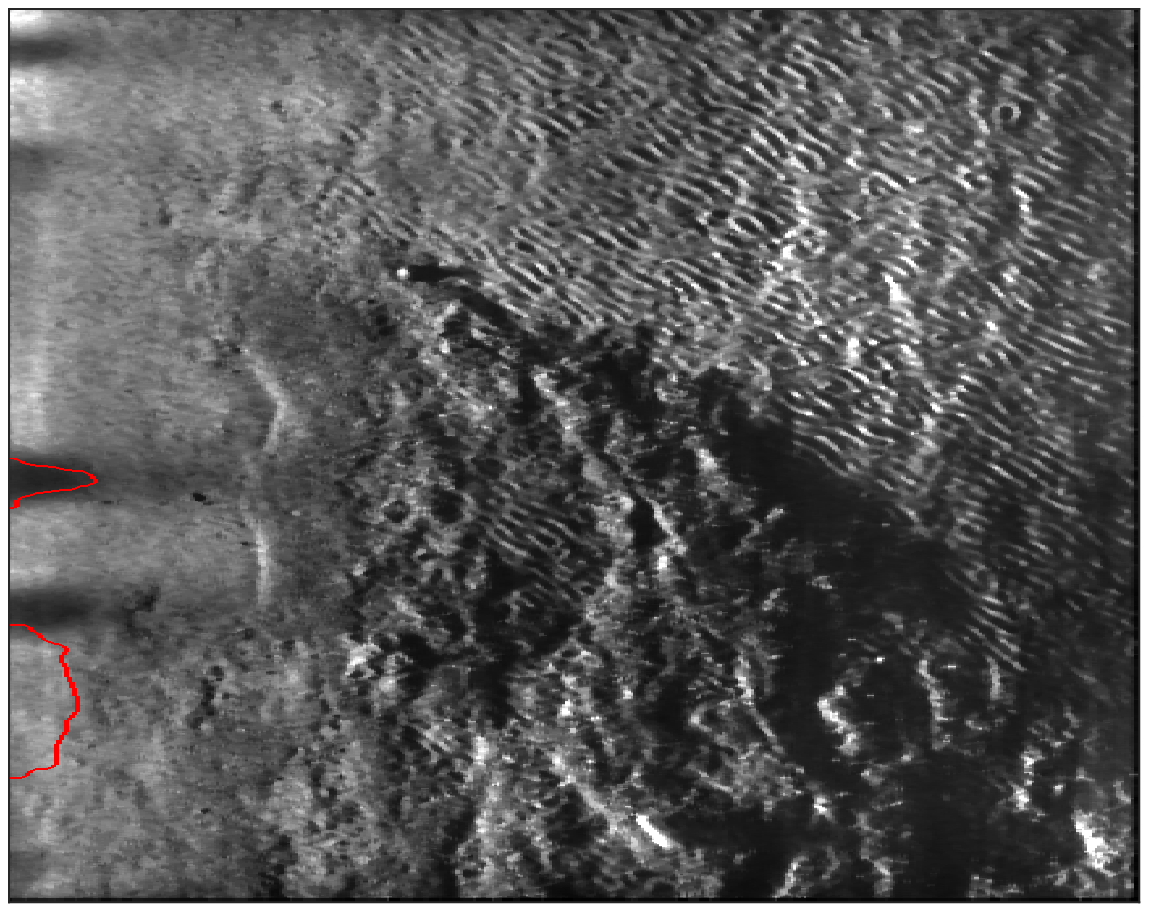}
	\caption{IoU}
\end{subfigure}
\begin{subfigure}[t]{.105\textwidth}
	\centering
	\includegraphics[trim={0cm 0cm 0cm 0cm},clip,width=\linewidth]{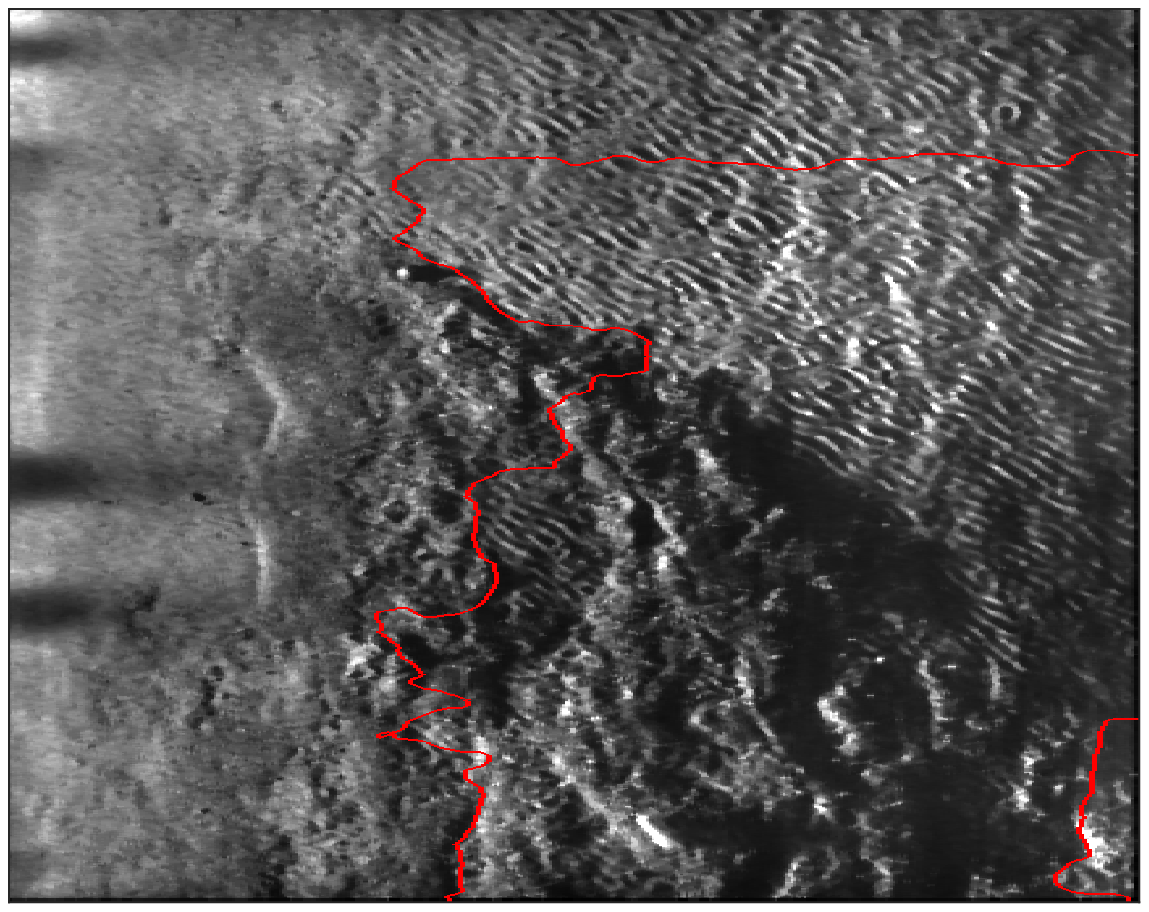}
	\caption{GCE}
\end{subfigure}
\begin{subfigure}[t]{.105\textwidth}
	\centering
	\includegraphics[trim={0cm 0cm 0cm 0cm},clip,width=\linewidth]{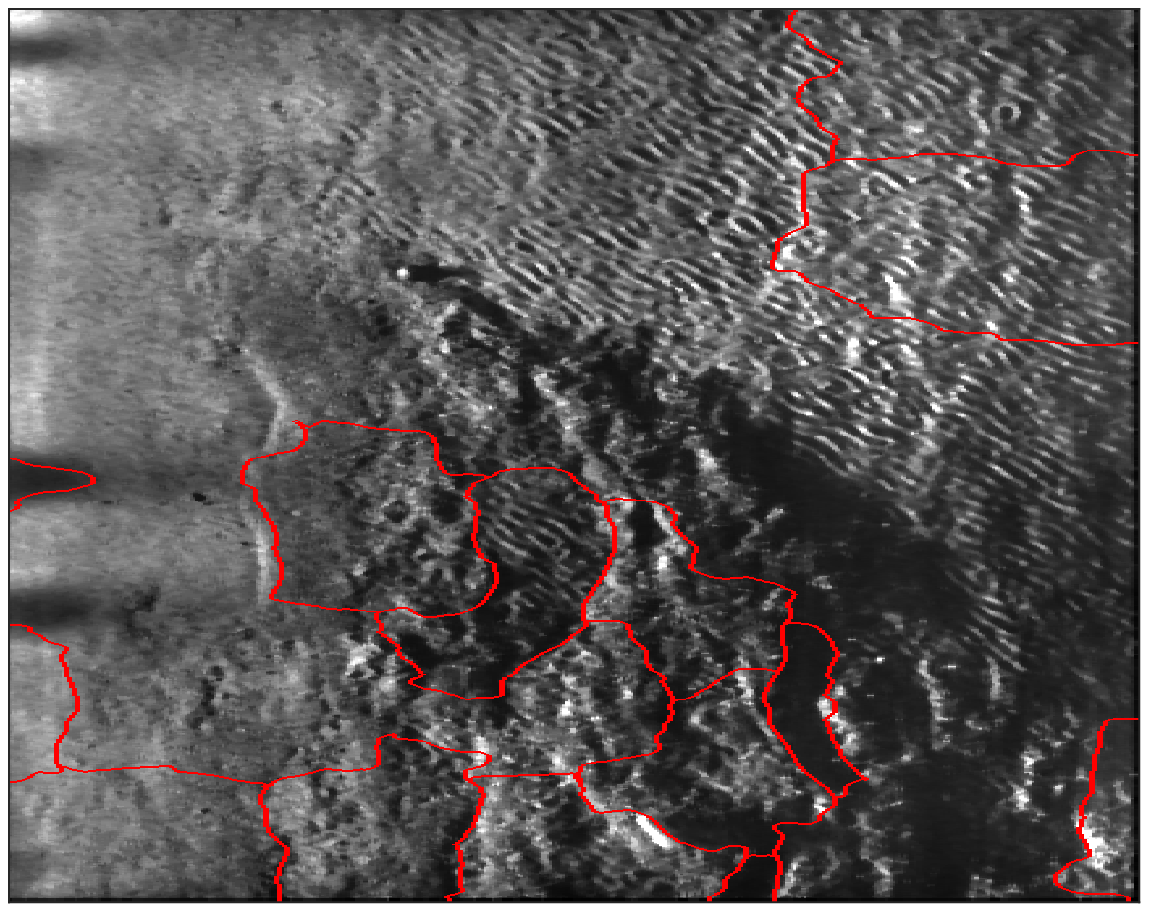}
	\caption{RI}
\end{subfigure}
\begin{subfigure}[t]{.105\textwidth}
	\centering
	\includegraphics[trim={0cm 0cm 0cm 0cm},clip,width=\linewidth]{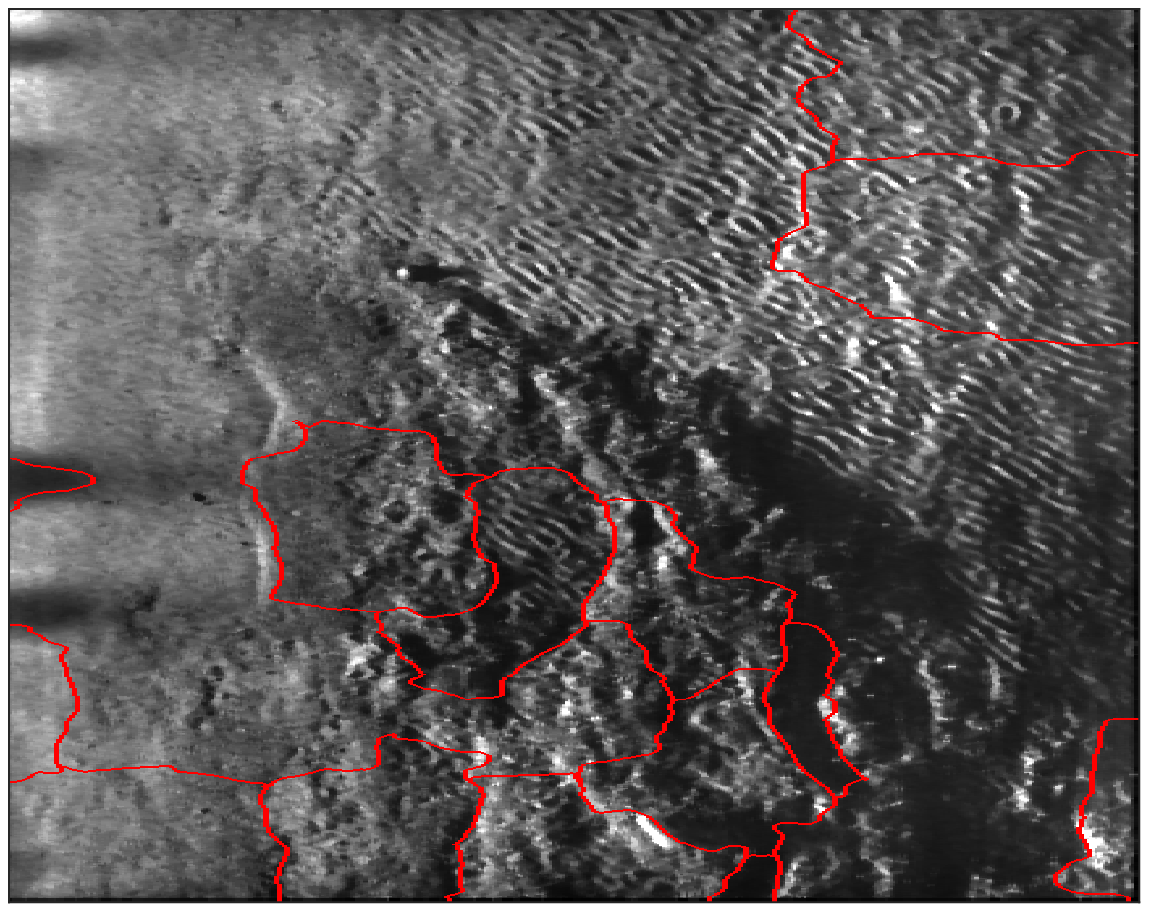}
	\caption{AR}
\end{subfigure}
\begin{subfigure}[t]{.105\textwidth}
	\centering
	\includegraphics[trim={0cm 0cm 0cm 0cm},clip,width=\linewidth]{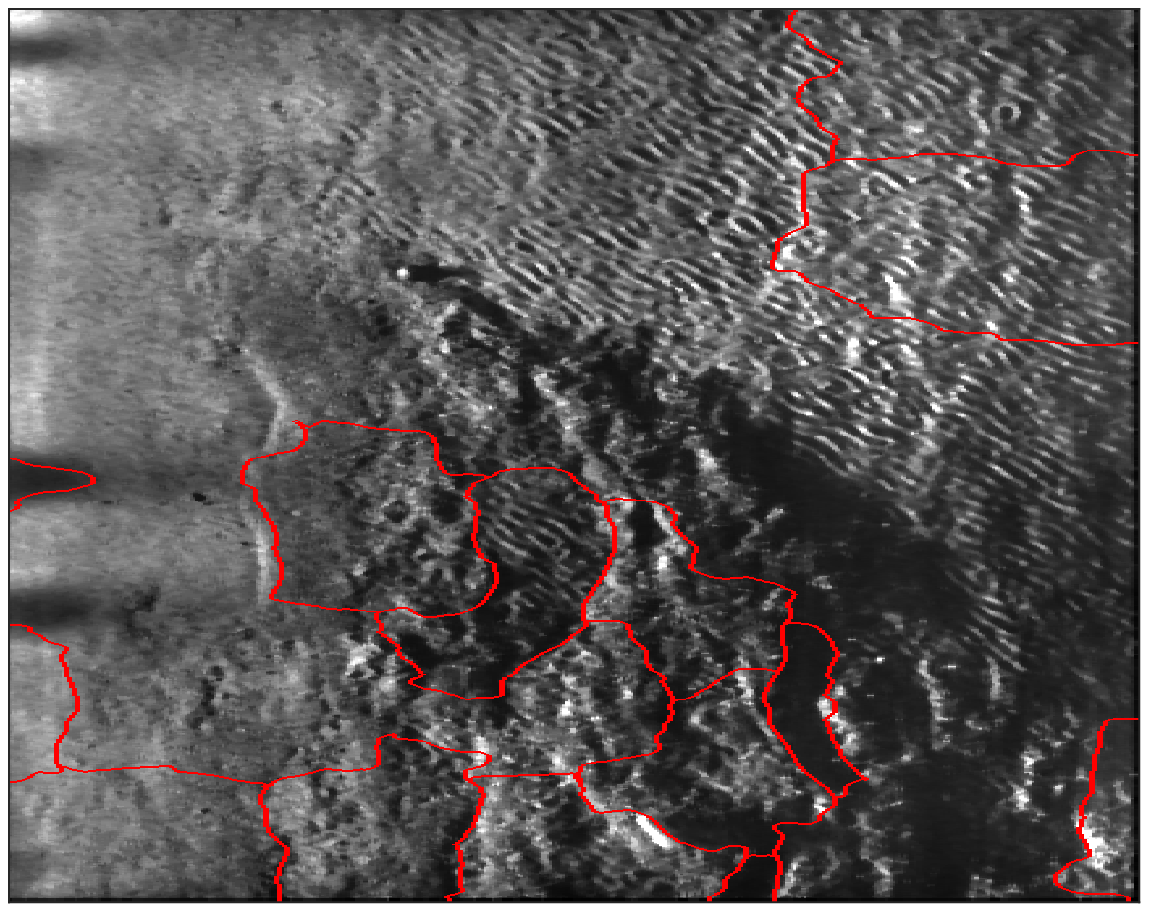}
	\caption{HI}
\end{subfigure}
\begin{subfigure}[t]{.105\textwidth}
	\centering
	\includegraphics[trim={0cm 0cm 0cm 0cm},clip,width=\linewidth]{Figures/Merging/WL-Rand/s2/PHi00-15Oct21_1246-00000006Edist.eps}
	\caption{Ours}
	\label{fig:need4}
\end{subfigure}
\caption{The best candidate selected by each quantitative metric for a given set of labels. Columns 1-3 of each row are the image, \textit{must-link} or crisp labels, and \textit{cannot-link} labels. Each row represents a single experiment comparing candidate segmentations with a unique set of labels. The columns right of the labels are the best result from segmentation indices followed by our approach. Our method uses the \textit{must-link} labels of column two and \textit{cannot-link} labels of column three. Every other quantitative metric relies on the crisp desired labels that are shown in column two. As the labels are reduced and changed down each row, selected candidates for each quantitative metric change except those selected by the WL-Rand. Slight changes in the labels modifies the performance of all indices except the WL-Rand. Additionally, when there is a significantly reduced amount of desired labels, as shown in the third row compared to the top row, all quantitative metrics select a candidate that mixes the rocky and sand-ripple texture except the WL-Rand.}
\label{fig:GoodMerge}
\end{figure*}

The initial superpixel segmentation is fixed and evaluation scores are computed for each candidate segmentation at every iteration of hierarchical merging. To compare our approach to the state-of-the-art, we provide an example of the best scoring candidate selected by each quantitative evaluation metric for the three sets of labels shown in Figure \ref{fig:GoodMerge}. The data was labeled with two sets of crisp labels in the first two rows and one set of weak-labels in the third row. For each experiment, all quantitative metrics except the WL-Rand use the labels in Figure \ref{fig:ML} as crisp labels, while the WL-Rand use the labels in Figure \ref{fig:ML} and Figure \ref{fig:CL}. The amount of labeled pixels is reduced from the first row with most of the image labeled to the last row with less than half of the pixels labeled. The IoU, GCE, RI, AR, and HI each change the selected partition when the amount and orientation of the crisp labels are slightly changed in each experiment. Meanwhile, regardless of slight changes to the labels, the best WL-Rand score chooses a consistent candidate. This demonstrates the importance of the \textit{cannot-link} labels and the penalty within the WL-Rand for crossing labeled boundaries. In the presence of few labels, as shown in the bottom row, the IoU, GCE, RI, AR, and HI each select a candidate partition that mixes the rocky and sand-ripple textures. However, the selected partition from the WL-Rand separates the sand-ripple in the top right of the image from the rocky portion of the image while the other remaining superpixels contain pure textures. This demonstrates the ability of our scheme to pick a good output for a further scene understanding algorithm. We also use WL-Rand to differentiate the comparison measures in the hierarchical merging algorithm. 

 The goal of merging superpixels is to have the fewest superpixels that represent pure texture regions. In practice, the number of desired superpixels apriori to hierarchical merging is unknown; therefore, we rely on the average of WL-Rand scores for every merge given a comparison function over a set of 37 textured SAS images as shown in Table \ref{tab:AverageSuccessScores}. Overall, using Earth Movers Distance \cite{ComparingDistRubner98} with CNN features performs the best. Comparison functions that utilize hand-crafted and CNN features performed better than those relying on simple feature representations \cite{cobb2014boundary,lacunarity}. While using Euclidean distance between convolutional feature vectors performs well (Euc), using the size of the superpixels in correspondence with representative features (EMD) provides a significant boost.

\begin{table}
	\centering
	\parbox[t][][t]{\linewidth}{\centering \caption{Average WL-Rand scores}
		\label{tab:AverageSuccessScores}
	\begin{tabular}{|l|c|l|c|}
		\hline
		Method  & WL-Rand & Method & WL-Rand \\
	\hline
	JS-Div \cite{cobb2014boundary} & $0.68 \pm 0.16$ &EMD-CNN & $\bm{0.77 \pm 0.10}$  \\
	\hline
	Lac  \cite{lacunarity} &$0.69\pm 0.17$ &Euc-CNN &$0.70 \pm 0.14$ \\
	\hline
	Mean-HC & $0.57 \pm 0.09$&Mean-CNN & $0.54 \pm 0.10$ \\
	\hline
	Edist-HC \cite{RizzoHierarchical}&$0.73 \pm 0.17$&Edist-CNN & $0.68 \pm 0.14$    \\ 
	\hline
	MI-HC\cite{renyiGonzalo} &$0.57 \pm 0.10$&MI-CNN &$ 0.68 \pm 0.13$ \\
	\hline
	\end{tabular}
}
\end{table}

\section{Summary}
\label{conclusions}
This paper introduces the Weakly-Labeled Rand index, a novel adaptation of the Rand index for weakly-labeled data. The developed WL-Rand metric aims to score a segmentation in relation to a weakly-labeled desired partition. The RI has terms that are analogous to the correct and incorrect \textit{must-link} and \textit{cannot-link} groups. If the \textit{must-link} and \textit{cannot-link} groups are interchangeable and the importance of few segments is equally as important as adhering to strict boundaries, then the WL-Rand and Rand are equivalent. The main advantage of using the WL-Rand is that in many remote sensing frameworks the desired partitioning is not crisp and we may only have knowledge of few weak labels.

We compare to state-of-the-art segmentation indices and display superior qualitative performance. Stability of the WL-Rand approach with regards to small changes in the desired labels is supported from the qualitative and quantitative comparisons to the traditional measures. We also use our WL-Rand index to compare several high-dimensional distance metrics for region merging. Out of all comparison functions, the EMD measure with pre-trained convolutional features performs the best in comparing neighboring segments. 
\bibliographystyle{ieeetr}
\bibliography{HierarchicalClusteringPaper}

\end{document}